\documentclass[sigconf]{acmart}
\usepackage{soul, color, xcolor}
\usepackage{multirow}
\usepackage{makecell}
\usepackage[table]{xcolor}
\usepackage[most]{tcolorbox}
\usepackage{utfsym}
\usepackage{enumitem}
\usepackage{color}
\usepackage{marvosym}

\makeatletter
\def\@fnsymbol#1{%
  \ifcase#1\or
    \Letter
  \or
    \dagger
  \or
    \ddagger
  \or
    \mathsection
  \or
    \mathparagraph
  \or
    \|
  \or
    **
  \or
    \dagger\dagger
  \or
    \ddagger\ddagger
  \else
    \@ctrerr
  \fi}
\makeatother

\definecolor{backred}{RGB}{255, 190, 190}
\definecolor{backblue}{RGB}{210, 230, 250}
\definecolor{darkgreen}{RGB}{1,50,32}

\newcommand{\best}[1]{\colorbox{backred!75}{\textbf{#1}}}
\newcommand{\second}[1]{\colorbox{backblue!75}{\textbf{#1}}}

\AtBeginDocument{%
  }

\copyrightyear{2026}
\acmYear{2026}
\setcopyright{cc}
\setcctype{by}
\acmConference[SIGIR '26] {Proceedings of the 49th International ACM SIGIR Conference on Research and Development in Information Retrieval}{July 20--24, 2026}{Melbourne, VIC, Australia.}
\acmBooktitle{Proceedings of the 49th International ACM SIGIR Conference on Research and Development in Information Retrieval (SIGIR '26), July 20--24, 2026, Melbourne, VIC, Australia}
\acmISBN{979-8-4007-2599-9/2026/07}
\acmDOI{10.1145/3805712.3809680}
\begin{document}

\title[Leveraging Multimodal Knowledge Graphs in Retrieval-Augmented Generation for Knowledge-intensive VQA]{mKG-RAG: Leveraging Multimodal Knowledge Graphs in Retrieval-Augmented Generation for Knowledge-intensive VQA}



\author{Xu Yuan}
\email{xuyuan127@gmail.com}
\affiliation{%
  \institution{The Hong Kong Polytechnic University}
  \state{Hong Kong}
  \country{China}
}

\author{Liangbo Ning}
\email{BigLemon1123@gmail.com}
\affiliation{%
  \institution{The Hong Kong Polytechnic University}
  \state{Hong Kong}
  \country{China}
}

\author{Qingqing Ye}
\email{qqing.ye@polyu.edu.hk}
\affiliation{%
  \institution{The Hong Kong Polytechnic University}
  \state{Hong Kong}
  \country{China}
}

\author{Wenqi Fan}
\authornote{Corresponding Author: Wenqi Fan.}
\email{wenqifan03@gmail.com}
\affiliation{%
  \institution{The Hong Kong Polytechnic University}
  \state{Hong Kong}
  \country{China}
}

\author{Qing Li}
\email{qing-prof.li@polyu.edu.hk}
\affiliation{%
  \institution{The Hong Kong Polytechnic University}
  \state{Hong Kong}
  \country{China}
}


\begin{abstract}
Retrieval-Augmented Generation (RAG) has emerged as an effective paradigm for expanding the knowledge capacity of Multimodal Large Language Models (MLLMs) by incorporating external knowledge sources into the generation process, and has been widely adopted for \textbf{knowledge-based Visual Question Answering (VQA)}.
Despite impressive advancements, vanilla RAG-based VQA methods that rely on unstructured documents and overlook the structural relations among knowledge elements frequently introduce irrelevant or misleading content, degrading answer accuracy and reliability. 
To overcome these challenges, a promising solution is to integrate \textbf{multimodal knowledge graphs (KGs)} into RAG-based VQA frameworks, thereby enhancing generation through structured multimodal knowledge.
To this end, this paper proposes \textbf{mKG-RAG}, a novel retrieval-augmented generation framework built upon multimodal KGs for knowledge-intensive VQA tasks.
Specifically, mKG-RAG leverages MLLM-driven graph extraction and vision-text matching to distill semantically consistent, modality-complementary entities and relations from multimodal documents, constructing high-quality multimodal KGs as structured knowledge representations.
Furthermore, a dual-stage retrieval strategy equipped with a query-aware multimodal retriever is introduced to improve retrieval efficiency while progressively refining precision.
Comprehensive experiments demonstrate that our approach significantly outperforms existing approaches and sets new state-of-the-art results for knowledge-based VQA. The code is available at \href{https://github.com/xandery-geek/mKG-RAG}{https://github.com/xandery-geek/mKG-RAG}.
\end{abstract}

\ccsdesc[500]{Information systems~Information retrieval}

\keywords{Retrieval-Augmented Generation, Multimodal Knowledge Graph, Visual Question Answering, Multimodal Large Language Model.}


\maketitle

\section{Introduction}
\label{sec:intro}

\begin{figure}[t]
    \centering
    \includegraphics[width=\linewidth]{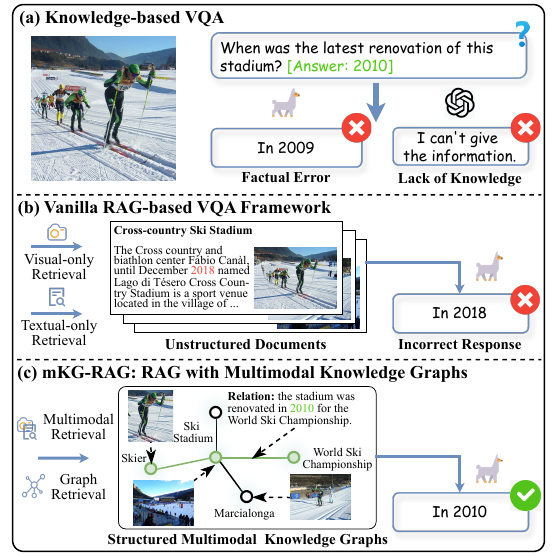}
    \caption{(a) Illustration of issues in knowledge-based VQA. (b) Vanilla RAG methods often underperform because they retrieve unstructured external knowledge using unimodal retrievers. (c) Our mKG-RAG augments MLLMs with structural information from multimodal knowledge graphs.}
    \label{fig:motivation}
\end{figure}

Visual Question Answering (VQA)~\cite{antol2015vqa,goyal2017making} is a challenging task at the intersection of vision and language, requiring models to jointly interpret images and questions to produce accurate answers.
This capability has supported applications in various domains, including medical image diagnosis~\cite{lin2023medical} and vision navigation~\cite{gordon2018iqa}.
Benefiting from powerful vision-language understanding and reasoning capabilities, Multimodal Large Language Models (\textbf{MLLMs})~\cite{liu2024improved,wang2409qwen2,chen2024internvl,wu2024deepseek} have provided a promising solution to conventional VQA tasks. 
For instance, LLaVA~\cite{liu2024improved} demonstrates strong zero-shot performance on commonsense VQA by integrating a pre-trained visual encoder for image representation with a large language model (LLM)~\cite{touvron2023llama} for answer reasoning.
Despite notable advancements, MLLMs face critical limitations in \textbf{knowledge-based VQA} scenarios~\cite{mensink2023encyclopedic,chen2023can}, also referred to as knowledge-intensive VQA in prior work, particularly those requiring encyclopedic knowledge, long-tail facts, or contextual reasoning beyond immediate visual inputs. 
As illustrated in Figure~\ref{fig:motivation} (a), when queried about the latest renovation date of a stadium, MLLMs exhibit two characteristic failure modes: generating plausible but factually incorrect responses or refusing to answer altogether. These issues stem from the scarcity of relevant knowledge in MLLMs' training corpus and the inherent difficulty of memorizing low-frequency facts~\cite{chen2023can}.

Recently, Retrieval-Augmented Generation (\textbf{RAG})~\cite{fan2024survey} has shown great potential in alleviating these challenges by incorporating external knowledge to complement the parametric memory of MLLMs, thereby enabling more accurate answer generation~\cite{lin2023fine,caffagni2024wiki,cocchi2024augmenting}.
Specifically, multiple query-relevant documents are retrieved from external knowledge bases and serve as in-context evidence to guide the generation process of MLLMs.
Despite their empirical success, vanilla RAG-based VQA approaches that rely on unstructured documents or paragraphs often introduce unnecessary noise and even misleading information~\cite{ma2025thinkongraph,wang2025knowledge}, which substantially compromises answer accuracy and reliability.
Moreover, these approaches typically overlook the structural relations among knowledge elements, limiting the reasoning capabilities of MLLMs. 
As illustrated in Figure~\ref{fig:motivation} (b), the presence of noisy and unstructured context makes it difficult for models to identify and exploit truly relevant supporting facts.
To address these limitations, a natural direction is to retrieve and leverage structured knowledge sources, such as knowledge graphs (KGs)~\cite{hogan2021knowledge,dai2025how}, which offer compact and well-organized representations to augment generation~\cite{edge2024graphrag,guo2024lightrag}.
In the VQA setting, which inherently involves multimodal reasoning, relying solely on textual KGs is suboptimal, as both visual and textual modalities are crucial for identifying relevant knowledge. 
Consequently, integrating \textbf{multimodal knowledge graphs} into the retrieval-augmented VQA framework presents a more robust solution for generating reliable, precise responses in knowledge-intensive scenarios, as illustrated in Figure~\ref{fig:motivation} (c).

Nevertheless, effectively retrieving relevant information from multimodal KGs to support knowledge-based VQA remains highly challenging.
First, off-the-shelf multimodal KGs~\cite{liu2019mmkg} are generally built around common entities and often lack the encyclopedic or long-tail content required by knowledge-intensive questions, rendering them ineffective for direct application.
Moreover, knowledge sources commonly used in knowledge-based VQA~\cite{mensink2023encyclopedic,chen2023can} are typically organized in unstructured documents that contain substantial contextual noise, making it difficult to extract well-structured multimodal entities and relations for constructing high-quality multimodal KGs.
Furthermore, building large-scale KGs from millions of documents, each potentially involving hundreds of entities and relations, dramatically expands the search space.
As a result, direct retrieval over such graphs becomes computationally inefficient, adversely affecting retrieval precision.

To overcome the above challenges, we propose \textbf{mKG-RAG}, a novel retrieval-augmented generation framework that integrates multimodal knowledge graphs to enhance the reasoning capabilities of MLLMs for knowledge-based VQA.
At the core of mKG-RAG is a multimodal knowledge graph construction module that converts unstructured multimodal documents (e.g., Wikipedia pages) into structured knowledge representations. This module leverages MLLM-driven graph extraction, along with vision–text matching, to identify semantically consistent and modality-complementary entities and relations from external multimodal sources.
To support efficient retrieval, mKG-RAG further develops a dual-stage multimodal retrieval paradigm that combines document-level recall with multi-granularity graph retrieval. 
The first stage efficiently narrows the search space by recalling candidate documents likely to contain relevant information, while the second stage refines the evidence by retrieving query-relevant subgraphs from dynamically constructed multimodal KGs derived from these candidate documents. 
Moreover, unlike prior approaches that rely on isolated unimodal retrievers, mKG-RAG employs a query-aware multimodal retriever trained on a high-quality query-evidence dataset, which substantially improves evidence relevance across modalities.
Comprehensive experiments on two widely used benchmarks demonstrate the effectiveness of mKG-RAG, achieving accuracies of 36.3\% on E-VQA and 40.5\% on InfoSeek.

The contributions of this work are summarized as follows:
\begin{itemize}
    \item We formulate multimodal KG-augmented RAG to address the challenges of knowledge-intensive VQA. By integrating multimodal graph structures into RAG, our approach captures structural relations among knowledge elements while preserving complementary cues across modalities, resulting in more accurate and reliable responses.
    \item The proposed mKG-RAG introduces a multimodal KG construction pipeline that extracts image–text–aligned entities and relations from multimodal documents, along with a dual-stage retrieval scheme featuring a query-aware multimodal retriever for efficient and precise graph retrieval.
    \item Extensive experiments demonstrate that mKG-RAG significantly outperforms strong baselines, setting new state-of-the-art results on E-VQA and InfoSeek benchmarks.
\end{itemize}

\begin{figure*}[htbp]
    \centering
    \includegraphics[width=\linewidth]{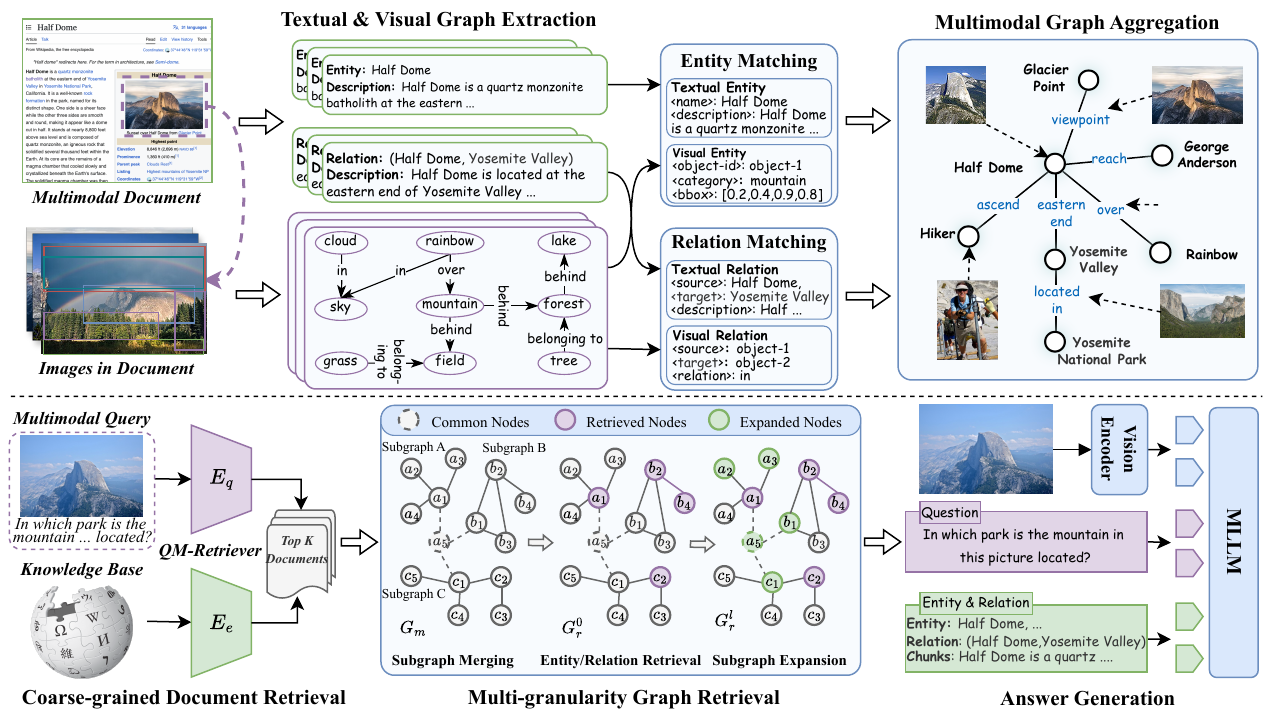}
    \caption{An overview of the proposed mKG-RAG framework, which comprises a multimodal knowledge graph construction pipeline (\textit{top}) and a dual-stage multimodal retrieval paradigm to facilitate answer generation (\textit{bottom}).}
    \label{fig:framework}
\end{figure*}

\section{Related Work}
\subsection{Knowledge-based VQA}
While traditional VQA~\cite{antol2015vqa,goyal2017making} benchmarks evaluate vision-language understanding primarily within the visual context, knowledge-intensive VQA significantly increases the challenge by requiring specific or detailed knowledge beyond the image content. 
Early benchmarks such as OK-VQA~\cite{marino2019ok} and A-OKVQA~\cite{schwenk2022okvqa} highlight the importance of commonsense knowledge in VQA, which can be effectively addressed by MLLMs trained on large and diverse corpora. 
However, E-VQA~\cite{mensink2023encyclopedic} and InfoSeek~\cite{chen2023can} introduced greater challenges by encompassing a broad range of Wikipedia entities and requiring fine-grained knowledge about them. 
Consequently, modern MLLMs often perform poorly on these benchmarks, as the relevant knowledge is missing or long-tailed in the training corpus~\cite{jiang2025qa}.

\subsection{Retrieval-Augmented Generation (RAG)}
RAG is commonly used in LLMs~\cite{brown2020language,ouyang2022training,jiang2025hibench,wang2025graph} to tackle issues such as outdated information and hallucinations~\cite{fan2024survey,liu2026inference,luo2025hv}. 
By dynamically combining external knowledge with the model's built-in capabilities, RAG provides an efficient solution for knowledge-intensive tasks~\cite{ning2025retrieval,jiang2026superglasses,huang2026rerec}.
Inspired by RAG, RA-VQA~\cite{lin2022retrieval}, RA-VQA-v2~\cite{lin2023fine}, and Wiki-LLaVA~\cite{caffagni2024wiki} have successfully applied retrieval augmentation to knowledge-intensive VQA, but their retrieval suffers from the modality gap~\cite{liang2022mind,zhang2024badcm,he2026cross,huang2026grading} between multimodal query and textual knowledge base. 
Subsequently, EchoSight~\cite{yan2024echosight} sequentially processes visual retrieval and multimodal re-ranking. However, EchoSight overlooks textual cues during retrieval and fails to incorporate visual information from the knowledge base during re-ranking.
Recent studies~\cite{cocchi2024augmenting,zhang2024mr} leveraged MLLMs to identify relevant information from retrieved passages, but this approach requires multiple MLLM calls and introduces substantial inference overhead.
Furthermore, existing RAG-based methods typically retrieve unstructured documents, overlooking both the noise in retrieval sources and the logical relations among knowledge elements, thereby yielding disorganized knowledge and increasing the reasoning load on MLLMs.

\subsection{Graph-based RAG}
Recently, the research community has begun exploring the use of Knowledge Graphs (KGs)~\cite{hogan2021knowledge,fan2019graph,wang2020traffic}, structured representations of entities and their relations, to enhance the reasoning and generation capabilities of LLMs~\cite{luo2024reasoning,edge2024graphrag,ma2025thinkongraph}.
For instance, GraphRAG~\cite{edge2024graphrag} automatically constructs textual KGs from documents using LLMs and employs community detection to facilitate query-focused summarization. 
To overcome the limitations of purely textual KGs, recent studies, such as MMGraphRAG~\cite{wan2025mmgraphrag}, QDM-GraphRAG~\cite{bu2025query}, and RAG-Anything~\cite{guo2025rag}, have extended the framework to multimodal KGs by integrating information from diverse modalities, including text, images, and tables.
Despite these advances, most existing methods represent non-textual content only as entities, without explicitly capturing relations through multimodal data.
Moreover, they are primarily designed for textual QA rather than VQA and therefore struggle to bridge the modality gap between multimodal queries and the supporting evidence needed in VQA scenarios.
Another related work, MR-MKG~\cite{lee2024multimodal}, targets commonsense VQA and retrieves from off-the-shelf multimodal KGs~\cite{liu2019mmkg} containing common entities.
In contrast, we propose the first RAG framework specifically tailored for knowledge-intensive VQA. Our approach enables the construction of knowledge graphs composed of semantically consistent, modality-complementary entities and relations, and supports multi-granularity graph retrieval, thereby substantially improving knowledge-intensive answer generation.
\section{The Proposed Method: mKG-RAG}
In the knowledge-based VQA task, the model receives an image-question pair $(I_q, q)$ as input and is required to generate a textual answer $a$, with access to an external knowledge base $\mathcal{B}$ as additional context. 
In our setting, the knowledge source is composed of multimodal documents featuring both text articles $T$ and their corresponding image assets $I$, \textit{i.e.}, $\mathcal{B} = \{(T_i, I_i)\}_{i=1}^{N}$.
The core objectives of our multimodal retrieval-augmented generation framework are twofold: (1) to effectively convert the unstructured knowledge base $\mathcal{B}$ into structured multimodal KGs, and (2) to precisely retrieve query-relevant knowledge from multimodal KGs while capturing the underlying structural relations, thereby augmenting the knowledge scope of MLLMs.

Figure~\ref{fig:framework} illustrates the overall workflow of the proposed mKG-RAG framework, which incorporates two key innovations.
First, we introduce a multimodal knowledge graph construction pipeline that leverages MLLMs to transform unstructured multimodal documents into structured knowledge graphs.
Second, a dual-stage multimodal retrieval strategy is proposed that enables multi-granularity graph retrieval over a query-specific multimodal KG, dynamically aggregated from subgraphs of documents initially retrieved via embedding-based document retrieval.
To support effective evidence matching in this process, we further design a Query-aware Multimodal Retriever (\textbf{QM-Retriever}) that bridges the gap between multimodal queries and candidate evidence.

\subsection{Multimodal Knowledge Graph Construction}
Existing retrieval-augmented VQA models often suffer from noisy contexts and overlook structural relations, as they directly retrieve fragmented textual chunks. A promising alternative is to leverage structured knowledge sources, such as multimodal KGs.
However, off-the-shelf multimodal KGs~\cite{liu2019mmkg} are typically designed for common entities and are unsuitable for addressing VQA cases involving detailed or long-tail knowledge, let alone domain-specific or private knowledge.
Thus, this work explores an effective multimodal KG construction pipeline to extract semantically consistent, modality-complementary entities and relations from accessible multimodal documents.
Specifically, for each document $(T, I) \in \mathcal{B}$, where the article $T = \{t_{1}, ..., t_{n}\}$ typically contains multiple sections and $I = \{i_{1}, ..., i_{m}\}$ is a set of images, we first segment it into manageable pieces. 
Sections without images are split or merged based on a fixed chunk size~\cite{fan2024survey}, while sections with images are preserved in their entirety to maintain alignment between images and text.
As depicted in the top part of Figure~\ref{fig:framework}, each segment is then processed by three key modules. 
Textual Graph Extraction identifies entities and their relations from text, while Visual Graph Extraction detects prominent objects and their interactions from images. Finally, the Multimodal Graph Aggregation module fuses the textual and visual entities and relations into a unified multimodal graph.

\subsubsection{\textbf{Textual Graph Extraction.}}
Following prior work~\cite{guo2024lightrag}, we process each textual piece by prompting MLLMs to identify key entities (nodes) and meaningful relations (edges), thereby forming a textual subgraph $\mathcal{G}_{t} = (\mathcal{N}, \mathcal{E})$. 
As shown in Figure~\ref{fig:framework}, each entity $n_i \in \mathcal{N}$ contains a unique name and a detailed description, offering an abstract representation to facilitate subsequent retrieval. 
Each relation $e_{ij} \in \mathcal{E}$ connects head and tail entities $(n_i, n_j)$ and includes a concise summary of their relation.

\subsubsection{\textbf{Visual Graph Extraction.}}
The textual subgraph has distilled the skeleton of textual chunks, including informative entities and relations, but it lacks visual elements, a critical component in VQA tasks.
A naive strategy is to directly supply $\mathcal{G}_{t}$ with the corresponding images~\cite {liu2019mmkg}.
However, considering that images often contain multiple objects and background noise, we propose augmenting the textual subgraph with fine-grained region-level information. 
Each region may represent an individual entity or a relation involving two or more entities, as depicted in Figure~\ref{fig:framework}.
For simplicity, this work focuses exclusively on binary relations, leaving the investigation of hyper-relations~\cite{luo2025hypergraphrag} for future research. 
Specifically, we employ Scene Graph Generation (SGG) techniques~\cite{im2024egtr} to extract a precise visual graph for each image in $I$. 
The visual graph is formalized as $\mathcal{G}_{v} = (\mathcal{V}, \mathcal{R})$, where $\mathcal{V}=\{v_i\}_{i=1}^{N_v}$ represents the set of visual objects with predicted category labels and bounding boxes, and $\mathcal{R}=\{r_{ij}\}_{i\ne j}$ denotes the visual relations between objects.
Unlike object detection~\cite{ren2015faster}, SGG offers additional relational information, facilitating effective vision-text relation 
matching in subsequent steps.

\subsubsection{\textbf{Multimodal Graph Aggregation.}}
The core contribution of the construction pipeline lies in integrating textual and visual graphs to form a semantically consistent, modality-complementary multimodal graph.
Directly aligning textual and visual entities/relations based on image-text similarity~\cite{radford2021learning} is limited to shallow or global alignment, lacking the capacity to capture fine-grained, context-aware correspondences.
Benefiting from the strong vision–language understanding capabilities of MLLMs~\cite{liu2024improved,bai2025qwen25vltechnicalreport}, this work leverages MLLMs as vision–text matchers to effectively identify semantically consistent entities/relations across visual and textual modalities.
Based on this insight, we design the following vision–text matching prompt:

\begin{tcolorbox}[
    colback=gray!10, 
    colframe=black,
    arc=2pt, auto outer arc,
    width=\linewidth,
    boxrule=0.5pt,
    left=1pt,
    right=1pt,
    top=0.5pt,
    bottom=0.5pt]
\textit{\textless Prefix Instruction\textgreater \ \textless IMAGE\textgreater \ [Textual Entities \& Relations] [Visual Entities \& Relations]}
\end{tcolorbox}

Here, \textit{\textless Prefix Instruction\textgreater} explains the input format of textual and visual graphs and guides the MLLMs to match textual and visual entities and relations. 
\textit{\textless IMAGE\textgreater} denotes the corresponding image of the visual graph and contains only the original image, without extra regions.
To enable MLLMs to comprehend graph structures, we convert both $\mathcal{G}_t$ and $\mathcal{G}_v$ into natural language format.
For $\mathcal{G}_t$, each entity and relation is expressed using its name and associated description.
The visual objects and relations in $\mathcal{G}_v$ are encoded as \textit{``\textless object-id\textgreater: \textless category\textgreater,\textless bbox\textgreater''} and \textit{``\textless relation-id\textgreater: \textless source-object\textgreater,\textless relation\textgreater,\textless target-object\textgreater''}, respectively. 
Importantly, visual objects include only the predicted category and normalized bounding box, from which MLLMs can locate the corresponding region within \textit{\textless IMAGE\textgreater}, without requiring actual regional images~\cite{yuan2024osprey,yuan2024instruction}. 
This design enables efficient inference by allowing simultaneous processing of all objects and relations in $\mathcal{G}_v$.
To ensure MLLMs follow the prefix instruction and produce the desired output, we further enhance their reasoning ability by providing several high-quality exemplars. The detailed vision–text matching prompt is provided in Figure~\ref{fig:vision-text-prompt-mini}.

The entire vision–text matching process is formalized as:
\begin{equation}
    \mathcal{M} = \{(n,v)_i\}_{i=1}^{N_e} \cup \{(e,r)_j\}_{j=1}^{N_r} = \mathcal{F}_{mllm} (I, \mathcal{G}_t, \mathcal{G}_{v}).
\end{equation}
Here, $\mathcal{M}$ denotes a set comprising $N_e$ matched entities and $N_r$ matched relations.
For each visual object $v$ or relation $r$, if it visually depicts a textual node $n$ or edge $e$, the pair $(n, v)$ or $(e, r)$ is regarded as a match.
We then treat the image region of $v$ or $r$ as an attribute of its textual counterpart $n$ or $e$, thereby constructing a multimodal subgraph $\mathcal{G}$, as illustrated in Figure~\ref{fig:framework}.
Specifically, $\mathcal{G}$ takes the textual graph $\mathcal{G}_t$ as its structural backbone and augments it with aligned visual elements from $\mathcal{G}_v$.
Since a visual relation $r$ involves two objects, we merge them into a single region by taking the union of their bounding boxes.
Unlike prior work~\cite{wan2025mmgraphrag,guo2025rag}, which represents images only as entities, our method explicitly models relations in the visual modality as first-class components.

Following the above procedure, we construct an image–text-aligned multimodal subgraph for each document segment. Subgraphs from all segments are then aggregated into a complete graph by merging identical nodes and edges.
Importantly, this merging is performed only among segments of the same document, ensuring that each document yields an independent multimodal KG.
During retrieval, relevant KGs from multiple documents are dynamically composed according to the retrieval results.
Since the construction process is query-independent, the entire pipeline can be executed offline, and each document requires processing only once.

\subsection{Dual-stage Multimodal Retrieval}
To fully leverage the constructed multimodal KGs, we further introduce a dual-stage retrieval framework inspired by human cognitive processes. 
When encountering unfamiliar multimodal queries, humans typically:
(1) filter relevant supporting evidence from vast external multimodal sources, then
(2) analyze and organize the extracted information into coherent structures for reasoning~\cite{zhu2025knowledge}.
Our framework accordingly implements coarse-grained document retrieval followed by multi-granularity graph retrieval.

\begin{figure}[t]
    \centering
    \includegraphics[width=0.98\linewidth]{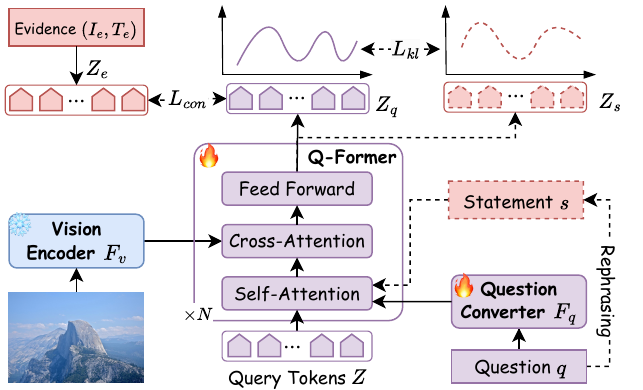}
    \caption{Architecture of Query-aware Multimodal Retriever}
    \label{fig:retriever}
\end{figure}

\subsubsection{\textbf{Coarse-grained Document Retrieval.}}
For a large-scale knowledge base containing millions of passages, direct graph retrieval is inefficient, as each passage may include hundreds of nodes and edges, greatly expanding the search space.
Thus, we first perform coarse-grained recall using vector search to identify candidate documents, as illustrated in the bottom part of Figure~\ref{fig:framework}.
Given a query $(I_q, q)$ and a set of multimodal documents $\{(T_i, I_i)\}_{i=1}^{N}$, a similarity set $S$ can be obtained:
\begin{equation}
    S = \left\{ s_i \mid s_i = \left\langle \mathcal{E}_q(I_q, q) \cdot \mathcal{E}_e(I_i, T_i) \right\rangle,\; i = 1, \ldots, N \right\},
\end{equation}
where $\langle \cdot \rangle$ denotes the cosine similarity. $\mathcal{E}_q$ and $\mathcal{E}_e$ denote the query and evidence encoders of the proposed QM-Retriever, which is designed to identify the most relevant multimodal evidence for a given image–question pair. 
A detailed description of the QM-Retriever is provided in Section~\ref{sec:qm_retriever}.
Based on the similarity set $S$, the top $K_d$ highest-scoring documents are collected.

\subsubsection{\textbf{Multi-granularity Graph Retrieval.}}
In contrast to prior chunk-based retrieval methods~\cite{yan2024echosight} that introduce substantial contextual noise, our approach retrieves evidence from compact, well-structured knowledge graphs distilled from candidate documents, enhancing retrieval precision and generation accuracy. 
Specifically, a query-specific multimodal graph $\mathcal{G}_m$ is constructed by online merging the offline-generated subgraphs corresponding to the candidate documents retrieved in the first stage. 
By restricting the merge to only query-specific documents, the online strategy effectively mitigates ambiguous entities and relations caused by cross-document knowledge inconsistencies~\cite{edge2024graphrag}.
Graph-based retrieval is then applied to identify query-relevant evidence by aggregating knowledge across entity, relation, and subgraph granularities.
First, embedding similarities between the multimodal query $(I_q, q)$ and each multimodal entity $(n, v)$ or relation $(e, r)$ in $\mathcal{G}_m$ are computed. The embedding vectors for the entity and relation are formalized as $f_e = \mathcal{E}_e(n, v)$ and $f_r = \mathcal{E}_e(e, r)$, respectively. 
Then, the top $K_g$ best-matched candidates are selected, \textit{e.g.}, entity $a_1$ and relation $(b_2, b_4)$ in Figure~\ref{fig:framework}. 
Combining these $K_g$ matched entities and relations produces an initial relevant subgraph $\mathcal{G}_r^0$.
Similarity-based retrieval alone may yield incomplete information, potentially omitting critical evidence needed to answer the question. 
To this end, we further exploit the structural properties of the graph~\cite{he2023structured,zhang2023tensorized} through subgraph expansion, incorporating information from $l$-hop neighbors of $\mathcal{G}_r^0$:
\begin{equation}
    \label{equ:graph-traversal}
    \mathcal{G}_r^{l} = \text{Graph Traversal} (\mathcal{G}_m, \mathcal{G}_r^{0}, l),     
\end{equation}
where $\mathcal{G}_r^{l}$ is the final relevant subgraph and $\text{Graph Traversal}$ is implemented via breadth-first search. 
Notably, we selectively incorporate only those neighbors whose embedding similarity to the query exceeds the threshold, as illustrated by the green nodes in Figure~\ref{fig:framework}.

The retrieved context comprises both graph elements (entities and relations) in $\mathcal{G}_r^{l}$ and their associated textual chunks. The former provides a structured knowledge outline, while the latter supplies the contextual details. 
Finally, the concatenated image, question, and context are fed into MLLMs for answer generation.

\subsubsection{\textbf{Query-aware Multimodal Retriever (QM-Retriever).}}
\label{sec:qm_retriever}
Standard multimodal retrievers prioritize semantic similarity over query relevance, often retrieving related content that lacks the evidence required for accurate answer generation.
To address this issue, this work proposes a query-aware multimodal retriever designed for evidence retrieval in VQA tasks. 

As illustrated in Figure~\ref{fig:retriever}, the QM-Retriever consists of a Vision Encoder $\mathcal{F}_v$, a Question Converter $\mathcal{F}_q$, and a Querying Transformer (Q-Former)~\cite{li2023blip}. 
We adopt the pre-trained vision encoder from BLIP-2~\cite{li2023blip} as $\mathcal{F}_v$ to extract image features.
The Question Converter is introduced to mitigate grammatical mismatches between interrogative questions and declarative evidence texts that can otherwise degrade retrieval performance. 
Instead of explicitly rewriting questions in the language space, $\mathcal{F}_q$ directly maps questions into declarative-style representations in the latent space.
Architecturally, $\mathcal{F}_q$ is composed of two linear layers with an intermediate ReLU activation, which transforms the original question embeddings into declarative representations before they are fed into the Q-Former.
Given an image–question pair $(I_q, q)$, QM-Retriever leverages the Q-Former to learn a compact set of query tokens $Z$ that aggregate multimodal information from both visual and textual inputs, \textit{i.e.},
\begin{equation}
Z_q = \text{Q-Former}(Z, \mathcal{F}_v(I_q), \mathcal{F}_q(q)).
\end{equation}
The resulting output $Z_q$ serves as the query representation, which is subsequently aggregated via mean pooling and used for vector-based retrieval.
When operating as the evidence encoder $E_e$, the QM-Retriever omits the Question Converter, since evidence texts are already expressed in declarative form.

To optimize the QM-Retriever, a query-evidence dataset is built based on the training set of E-VQA~\cite{mensink2023encyclopedic}, where each multimodal query $(I_q, q)$ is paired with its corresponding ground-truth evidence $(I_e, T_e)$. 
Here, $T_e$ represents evidence text, and $I_e$ refers to the associated image from the evidence section. For sections without visual content, blank images are used as placeholders.
The optimization process involves two key objectives: 
(1) \textbf{\textit{Question Reformulation.}} We leverage LLMs to convert the original question $q$ into a declarative statement $s$ that emphasizes the scene context. 
By encoding $(I_q, s)$ with the QM-Retriever, we obtain a declarative representation $Z_{s}$ as a reference. Then, the Kullback-Leibler divergence is measured to minimize the divergence between the distributions of $Z_q$ and $Z_s$:
\begin{equation}
    \mathcal{L}_{kl} = \ D_{KL}(p(Z_q|I_q, q) \ \| \ p(Z_s|I_q, s)). 
\end{equation}
(2) \textbf{\textit{Question-Evidence Alignment.}} To retrieve query-relevant evidence, we employ contrastive learning~\cite{he2020momentum} to align the features of multimodal query and evidence by encouraging positive query-evidence pairs to have more similar representations than negative pairs in a batch, \textit{i.e.},
\begin{equation}
    \mathcal{L}_{con} = - \log \frac{\exp (\text{sim}(Z_q, Z_e) / \tau)}{\Sigma_{k=1}^{B} \exp(\text{sim}(Z_q, Z_k) / \tau)}.
\end{equation}
Here, $B$ denotes the batch size, and $\tau$ is a temperature parameter. 
Finally, the total objective is formulated as the linear combination controlled by a hyperparameter $\alpha$: $\mathcal{L} = \mathcal{L}_{con} + \alpha \mathcal{L}_{kl}$. 
Notably, the Q-Former is initialized with BLIP-2's weights and fine-tuned jointly with $\mathcal{F}_q$, while $\mathcal{F}_v$ remains frozen.

\begin{table*}[t]
\centering
\caption{VQA accuracy on the E-VQA and InfoSeek datasets. The \best{best performance} and the \second{second-best performance} are marked in red and blue, respectively. 
* denotes that the model is further fine-tuned on the corresponding dataset. In the Retrieval Mode column, ``Text'' and ``Vision'' indicate whether the retrieval query is based on textual or visual information. For brevity, \dag{} and \ddag{} denote the text-only and vision-only variants of the same model, respectively.
}
\label{tab:main-results}
\resizebox{\linewidth}{!}{
\begin{tabular}{lccccccccc}
\toprule
\multirow{2}{*}{\textbf{Model}} 
& \multirow{2}{*}{\textbf{LLM / MLLM}}
& \multicolumn{3}{c}{\textbf{Retrieval Mode}}
& \multicolumn{2}{c}{\textbf{E-VQA}} 
& \multicolumn{3}{c}{\textbf{InfoSeek}} \\
\cmidrule(lr){3-5} \cmidrule(lr){6-7} \cmidrule(lr){8-10}
&
& Retriever & Text & Vision
& Single-Hop & All  
& Unseen-Q & Unseen-E & All  \\
\midrule
\rowcolor{gray!20} \multicolumn{10}{c}{\textit{Zero-shot MLLMs}}                                                               \\
BLIP-2~\cite{li2023blip}                                  & Flan-T5XL    & --            & \usym{2717} & \usym{2717}  & 12.6       & 12.4 & 12.7     & 12.3     & 12.5 \\
LLaVA-v1.5~\cite{liu2024improved}                         & Vicuna-7B    & --            & \usym{2717} & \usym{2717}  & 16.3       & 16.9 & 9.6      & 9.4      & 9.5  \\
LLaVA-MORE~\cite{cocchi2025llava}                         & LLaMA-3.1-8B & --            & \usym{2717} & \usym{2717}  & 15.8       & 16.0 & 9.0      & 8.2      & 8.6 \\
Qwen2-VL~\cite{wang2409qwen2}                             & Qwen2-VL-7B & --             & \usym{2717} & \usym{2717}  & 19.9       & 19.7 & 19.8     & 18.5     & 19.2 \\
\midrule
\rowcolor{gray!20} \multicolumn{10}{c}{\textit{Retrieval-Augmented Models}}                                                    \\
RA-VQA\textsuperscript{*}~\cite{lin2022retrieval}         & T5-large     & BERT-base     & \usym{2713} & \usym{2717}  & 21.7       & 20.0 & 26.1     & 25.8     & 25.9 \\
RA-VQA-v2\textsuperscript{*}~\cite{lin2023fine}            & T5-large   & ColBERT \& CLIP & \usym{2713} & \usym{2713}  & 22.4       & 21.5 & 27.8     & 27.2     & 27.5 \\
RORA-VLM~\cite{qi2024rora}                                & Vicuna-7B    & CLIP \& GS    & \usym{2713} & \usym{2713}  & --         & 20.3 & 25.1     & 27.3     & --   \\
Wiki-LLaVA\textsuperscript{*}~\cite{caffagni2024wiki}     & Vicuna-7B    & CLIP ViT-L/14 & \usym{2713} & \usym{2717}  & 21.8       & 26.4 & 30.1     & 27.8     & 28.9 \\
EchoSight\dag~\cite{yan2024echosight}                     & LLaMA-3.1-8B & EVA-CLIP-8B   & \usym{2713} & \usym{2717}  & 22.4       & 21.7 & 30.0     & 30.7     & 30.4 \\
EchoSight\ddag~\cite{yan2024echosight}                    & LLaMA-3.1-8B & EVA-CLIP-8B   & \usym{2717} & \usym{2713}  & 26.4       & 24.9 & 18.0     & 19.8     & 18.8 \\
mR$^2$AG\textsuperscript{*}~\cite{zhang2024mr}            & Vicuna-7B    & CLIP ViT-L/14 & \usym{2717} & \usym{2713}  
& --         
& --   
& \second{40.6}
& \best{39.8}     
& \second{40.2} \\
ReflectiVA\textsuperscript{*}\dag~\cite{cocchi2024augmenting}  & LLaMA-3.1-8B & EVA-CLIP-8B   & \usym{2713} & \usym{2717}  & 28.0       & 29.2 & 40.4     
& \best{39.8}     & 40.1 \\
ReflectiVA\textsuperscript{*}\ddag~\cite{cocchi2024augmenting} & LLaMA-3.1-8B & EVA-CLIP-8B   & \usym{2717} & \usym{2713}  
& 35.5       & \second{35.5} & 28.6     & 28.1     & 28.3 \\
\midrule
\rowcolor{gray!20} \multicolumn{10}{c}{\textit{Graph Retrieval-Augmented Models}} \\
RAG-Anything~\cite{guo2025rag}                                              & LLaMA-3.1-8B & text-embedding-3-small & \usym{2713} & \usym{2717}  & 25.6       & 24.8 & 28.7     & 28.1     & 28.4 \\
\midrule
mKG-RAG\dag                                               & LLaMA-3.1-8B & CLIP ViT-L/14 & \usym{2713} & \usym{2717}  & 24.4       & 23.4 & 24.1     & 22.3     & 23.2 \\
mKG-RAG\ddag                                              & LLaMA-3.1-8B & CLIP ViT-L/14 & \usym{2717} & \usym{2713}  & 24.6       & 23.7 & 21.3     & 19.8     & 20.6 \\
mKG-RAG                                                   & LLaMA-3.1-8B & QM-Retriever  & \usym{2713} & \usym{2713}  & 27.1       & 26.1 & 32.9     & 31.3     & 32.1 \\
\midrule
mKG-RAG\textsuperscript{*}\dag                            & LLaMA-3.1-8B & CLIP ViT-L/14 & \usym{2713} & \usym{2717}  
& \second{36.6}       & 34.9 & 29.8     & 28.5     & 29.1 \\
mKG-RAG\textsuperscript{*}\ddag                           & LLaMA-3.1-8B & CLIP ViT-L/14 & \usym{2717} & \usym{2713}  & 32.9       & 31.0 & 29.4     & 27.3     & 28.3 \\
mKG-RAG\textsuperscript{*}                                & LLaMA-3.1-8B & QM-Retriever  & \usym{2713} & \usym{2713}  
& \best{38.4}       
& \best{36.3} 
& \best{41.4}     
& \second{39.6}     
& \best{40.5}       \\
\bottomrule
\end{tabular}
}
\end{table*}

\section{Experiments}
\subsection{Experimental Setup}
\noindent \textbf{Datasets.}
Experiments are conducted on the E-VQA~\cite{mensink2023encyclopedic} and InfoSeek~\cite{chen2023can} datasets, both of which comprise question–answer pairs grounded in Wikipedia documents.
The \textbf{E-VQA} test set contains 5.8K samples spanning \textit{Single-Hop} and \textit{Two-Hop} questions, where the former can be answered from a single Wikipedia page, whereas the latter requires sequential retrieval across multiple documents.
Answer correctness is evaluated using the BERT Matching (BEM) metric~\cite{bulian2022tomayto}, which measures the percentage of matches between predicted and ground-truth answers.
For \textbf{InfoSeek}, following prior work~\cite{yan2024echosight,cocchi2024augmenting}, we report results on the 73K-sample validation set, which includes \textit{Unseen-Q} (unseen questions) and \textit{Unseen-E} (unseen entities) subsets.
We adopt the dataset's evaluation metrics: VQA Accuracy~\cite{goyal2017making} for STRING and TIME questions, and Relaxed Accuracy~\cite{methani2020plotqa} for NUMERICAL questions.

\noindent \textbf{Knowledge Base.}
E-VQA provides a knowledge base comprising 2M Wikipedia pages, with each question-answer pair annotated with supporting Wikipedia articles, relevant evidence paragraphs, and associated images. 
For InfoSeek, since there is no publicly released knowledge base, we utilize a subset of 100K documents from E-VQA filtered by EchoSight~\cite{yan2024echosight}.

\noindent \textbf{Baselines.}
To evaluate the effectiveness of mKG-RAG, we compare it with various MLLMs (zero-shot) and several state-of-the-art RAG methods on knowledge-based VQA, as summarized in Table~\ref{tab:main-results}. Notably, the results of RA-VQA, RA-VQA-v2, and RAG-Anything are obtained from our reproduced implementations.

\noindent \textbf{Implementation Details.} 
We adopt Llama-3.2-11B-Vision~\cite{llama-3.2-vision} as the MLLM for multimodal KG construction, supporting both textual graph recognition and vision–text matching. For visual graph generation, we employ a lightweight one-stage SGG model, EGTR~\cite{im2024egtr}.
The QM-Retriever is trained for 25 epochs on 221K query–evidence pairs from the E-VQA training set using the AdamW optimizer with a CosineLR scheduler and an initial learning rate of $10^{-5}$. Training uses a batch size of 64, a KL divergence coefficient of 2, an input image resolution of $224 \times 224$, and a maximum token length of 512 for both questions and evidence.
For first-stage retrieval, embeddings produced by the QM-Retriever are indexed with FAISS~\cite{johnson2019billion} to enable efficient approximate nearest neighbor search, returning the top $K_d=10$ candidate documents. In graph retrieval, we empirically set $K_g=10$ and $l=1$.
Following ReflectiVA~\cite{cocchi2024augmenting}, we use LLaVA-MORE~\cite{cocchi2025llava} as the default multimodal answer generator. As ReflectiVA is optimized for passage filtering and answer generation, we implement a fine-tuned variant, mKG-RAG\textsuperscript{*}, trained on the E-VQA and InfoSeek splits together with LLaVA-Instruct data~\cite{liu2024improved}. Parameter-efficient tuning is performed using LoRA adapters~\cite{hu2022lora} with a total batch size of 32 and a learning rate of $1.5 \times 10^{-4}$.

\subsection{Performance Comparison}

\subsubsection{\textbf{Results on Knowledge-based VQA}}
In this section, we compare mKG-RAG with Zero-shot MLLMs and RAG-based approaches on the benchmarks mentioned above.
The results in Table~\ref{tab:main-results} demonstrate that Zero-shot MLLMs struggle with knowledge-based VQA tasks, particularly on the InfoSeek dataset. 
These limitations underscore the critical need for integrating external knowledge. By augmenting LLaVA-MORE with mKG-RAG, we achieve substantial improvements, over 20.3\% on E-VQA and 31.9\% on InfoSeek, highlighting the value of retrieval augmentation.

Furthermore, our method achieves the highest overall performance under the ``All'' set on both datasets. 
In the fine-tuning setting, mKG-RAG\textsuperscript{*} consistently outperforms mR$^2$AG\textsuperscript{*} and ReflectiVA\textsuperscript{*}, except on the ``Unseen-E'' subset of InfoSeek. 
Notably, even without fine-tuning, mKG-RAG still surpasses EchoSight by 1.2\% and 1.7\% on the two datasets, respectively. 
These results underscore the advantages of integrating RAG with multimodal KGs and validate the effectiveness of our QM-Retriever.
Compared to the graph-based RAG baseline RAG-Anything, mKG-RAG achieves additional gains of 1.3\% and 3.7\%. 
We attribute this to the fact that RAG-Anything, which largely converts multimodal data into textual KGs, inevitably loses modality-specific information, whereas our multimodal KGs exploit the complementary cues across modalities, leading to more accurate retrieval and reasoning.

Table~\ref{tab:main-results} also reports two mKG-RAG variants where the QM-Retriever is replaced by text-only and vision-only CLIP for entity and relation retrieval, while document retrieval for multimodal knowledge graph construction still relies on the QM-Retriever. 
The text-only variant uses both questions and image captions as queries, providing richer context and thus outperforming the vision-only variant. 
While mR$^2$AG\textsuperscript{*} and ReflectiVA\textsuperscript{*} perform competitively in unimodal settings due to their specialized designs, mKG-RAG\textsuperscript{*} is explicitly tailored for VQA with multimodal KGs and thus demonstrates clear advantages in multimodal retrieval, achieving over 10\% gains on InfoSeek compared to unimodal retrieval.

\begin{table*}[t]
\begin{minipage}[t]{0.49\textwidth}
\caption{Retrieval performance on E-VQA test set.}
\label{tab:retrieval-evaq}
\centering
\resizebox{\linewidth}{!}{
\begin{tabular}{ccccccc}
\toprule
\multirow{2}{*}{\textbf{Model}} 
& \multirow{2}{*}{\makecell[c]{\textbf{Retrieval} \\ \textbf{Mode}}} 
& \multicolumn{5}{c}{\textbf{E-VQA}}        \\
\cmidrule(lr){3-7}
                       &                            & R@1  & R@5  & R@10 & R@20 & R@50 \\
\midrule
Nomic-text             & T$\rightarrow$T            & 2.0  & 4.1  & 5.6  & 7.8  & 11.1 \\
Nomic-vision           & V$\rightarrow$V            & 9.3  & 23.0 & 29.3 & 36.0 & 45.6 \\
CLIP ViT-L/14          & T$\rightarrow$T            & 2.0  & 4.7  & 6.4  & 8.8  & 12.1 \\
CLIP ViT-L/14          & V$\rightarrow$V            
& \second{11.2} 
& \second{28.5} 
& \second{36.2} 
& \second{44.1} 
& \second{54.8} \\
\midrule
CLIP ViT-L/14          & T$\rightarrow$V            & 1.1  & 3.1  & 4.6  & 7.3  & 12.3 \\
CLIP ViT-L/14          & V$\rightarrow$T            & 3.8  & 10.2 & 13.6 & 18.0 & 23.9 \\
\midrule
QM-Retriever           & Multimodal                 
& \best{18.9} 
& \best{36.8} 
& \best{46.2} 
& \best{55.6} 
& \best{66.7} \\
\bottomrule
\end{tabular}
}
\end{minipage}
\hfill
\begin{minipage}[t]{0.49\textwidth}
\caption{Retrieval performance on InfoSeek validation set.}
\label{tab:retrieval-infoseek}
\centering
\resizebox{\linewidth}{!}{
\begin{tabular}{ccccccc}
\toprule
\multirow{2}{*}{\textbf{Model}} 
& \multirow{2}{*}{\makecell[c]{\textbf{Retrieval} \\ \textbf{Mode}}} 
& \multicolumn{5}{c}{\textbf{InfoSeek}}        \\
\cmidrule(lr){3-7}
                       &                            & R@1  & R@5  & R@10 & R@20 & R@50 \\
\midrule
Nomic-text             & T$\rightarrow$T            & 11.0 & 19.3 & 24.2 & 30.4 & 40.6 \\
Nomic-vision           & V$\rightarrow$V            & 35.0 & 56.5 & 63.3 & 69.3 & 75.5 \\
CLIP ViT-L/14          & T$\rightarrow$T            & 9.2  & 15.8 & 19.3 & 23.3 & 30.0 \\
CLIP ViT-L/14          & V$\rightarrow$V            
& \second{40.0} 
& \second{63.4} 
& \second{70.9} 
& \second{77.7} 
& \second{83.7} \\
\midrule
CLIP ViT-L/14          & T$\rightarrow$V            & 8.5  & 18.8 & 24.6 & 31.7 & 42.5 \\
CLIP ViT-L/14          & V$\rightarrow$T            & 20.1 & 40.1 & 49.2 & 58.3 & 68.9 \\
\midrule
QM-Retriever           & Multimodal                 
& \best{49.7} 
& \best{71.6} 
& \best{78.0} 
& \best{82.5} 
& \best{89.1} \\
\bottomrule
\end{tabular}
}
\end{minipage}
\end{table*}

\subsubsection{\textbf{Results on Retrieval}}
To assess the effectiveness of multimodal retrieval using the QM-Retriever, we compare it with unimodal and cross-modal retrievers in selecting the most relevant documents for VQA queries.
Specifically, we use Nomic-Embed-v1.5~\cite{nussbaum2025nomic} and CLIP ViT-L/14@336~\cite{radford2021learning} as retrieval baselines and examine four feasible retrieval combinations: text-to-text (T$\rightarrow$T), vision-to-vision (V$\rightarrow$V), text-to-vision (T$\rightarrow$V), and vision-to-text (V$\rightarrow$T).
Tables~\ref{tab:retrieval-evaq} and~\ref{tab:retrieval-infoseek} report the Recall scores on E-VQA and InfoSeek, respectively. The QM-Retriever consistently outperforms all baseline methods, achieving average improvements of 9.9\% (E-VQA) and 7.0\% (InfoSeek) over the second-best approach. The strong recall performance ensures that mKG-RAG operates on highly relevant KGs dynamically constructed in the multi-granularity graph retrieval phase, as further supported by our ablation studies.
Additionally, the results reveal that V$\rightarrow$V retrieval consistently outperforms other unimodal and cross-modal configurations, underscoring the critical role of visual content in VQA tasks. 
This finding validates our design choice to augment knowledge-based VQA with multimodal KGs rather than text-only alternatives.

\begin{table}[]
\centering
\caption{Performance of different mKG-RAG variants on single-hop and two-hop queries of E-VQA Benchmark.}
\label{tab:two-hop}
\resizebox{\linewidth}{!}{
\begin{tabular}{ccccc}
\toprule
\multirow{2}{*}{\textbf{Model}} & \multirow{2}{*}{\textbf{Retrieval Mode}} & \multicolumn{3}{c}{\textbf{E-VQA}}   \\
\cmidrule(lr){3-5}
                       &                                 & Single-Hop & Two-Hop & All  \\
\midrule
LLaVA-MORE                       & --                         & 15.8       & 17.1    & 16.0 \\
\midrule
mKG-RAG\textsuperscript{*}\dag   & Text-only                       & 36.6       & \best{26.7}    & 34.9 \\
mKG-RAG\textsuperscript{*}\ddag  & Vision-only                     & 32.9       & 21.5    & 31.0 \\
mKG-RAG\textsuperscript{*}       & Multimodal                      & \best{38.4}       & \best{26.7}    & \best{36.3} \\
\bottomrule
\end{tabular}
}
\end{table}
\subsubsection{\textbf{Performance on Two-Hop Questions}}
Table~\ref{tab:two-hop} presents the VQA accuracy of our approach on both single-hop and two-hop questions. The results show that mKG-RAG and its variants significantly outperform zero-shot LLaVa-More in two-hop scenarios, further validating the effectiveness of our method. 
However, the gains on two-hop questions are smaller than those on single-hop questions, likely because specialized techniques for multi-hop question-answering, such as question decomposition or iterative retrieval, are absent.

\subsubsection{\textbf{Computational Efficiency}}
Although mKG-RAG consists of multiple components, it remains computationally efficient because the most resource-intensive step, multimodal knowledge graph construction, is performed offline. 
The remaining online operations are summarized in Table~\ref{tab:efficiency}, where we compare mKG-RAG with two baselines: zero-shot LLaVA-MORE and Naive RAG, which replaces our dual-stage multimodal retrieval with simple chunk-based retrieval. 
We normalize all runtime costs by setting the cost of zero-shot generation to 1. Under this metric, mKG-RAG achieves a $1.37\times$ relative improvement in VQA accuracy over Naive RAG while incurring only a $1.14\times$ increase in total computational overhead, demonstrating a favorable trade-off between performance and efficiency.

\begin{figure*}
    \centering
    \includegraphics[width=\linewidth]{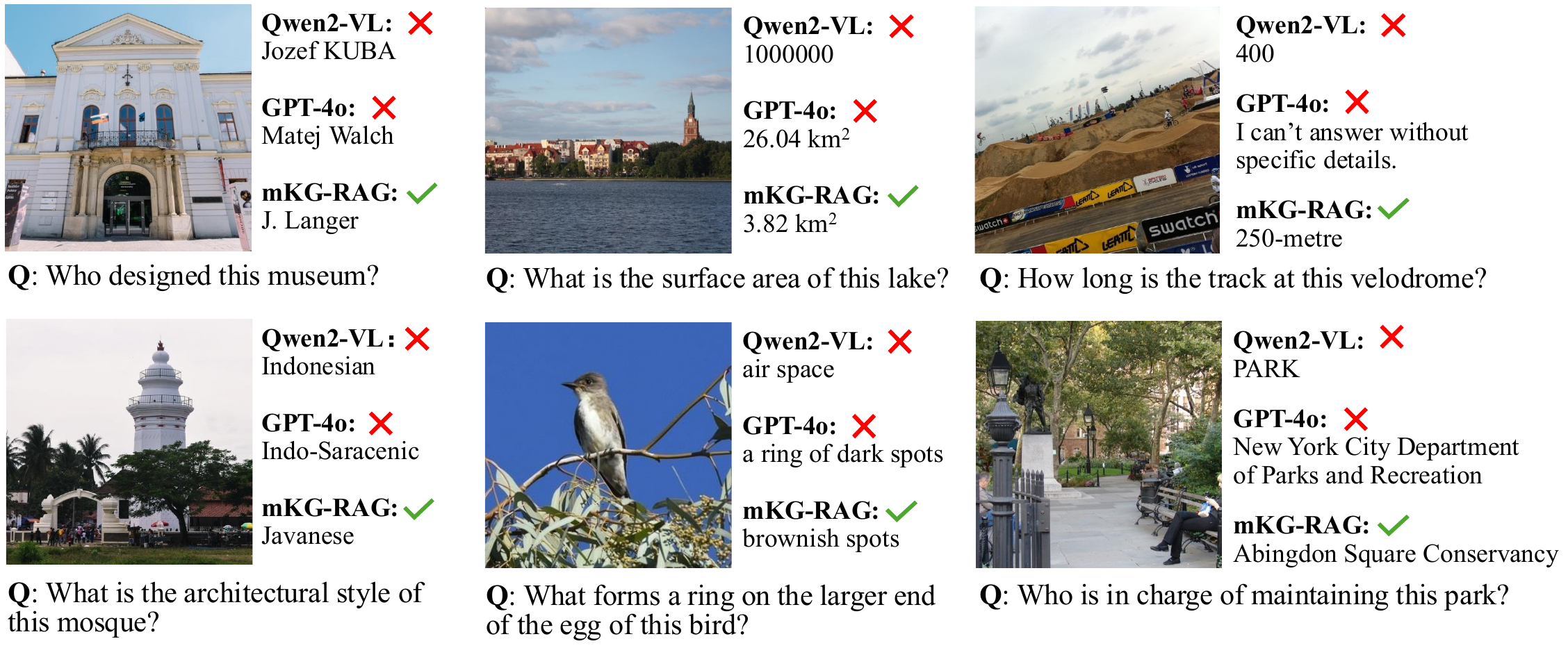}
    \caption{Qualitative results of Qwen2-VL-7B, GPT-4o, and mKG-RAG on the E-VQA dataset}
    \label{fig:visualization}
\end{figure*}

\subsubsection{\textbf{Qualitative Results}}
Figure~\ref{fig:visualization} presents a qualitative comparison between mKG-RAG and the zero-shot baselines Qwen2-VL-7B and GPT-4o. 
In these examples, Qwen2-VL-7B and GPT-4o frequently produce plausible yet incorrect responses or occasionally refuse to respond, illustrating trustworthiness limitations of current MLLMs in knowledge-intensive VQA settings. In contrast, mKG-RAG, augmented with multimodal knowledge graphs, consistently handles knowledge-intensive queries, particularly those requiring precise numerical answers.

\begin{table}[t]
\caption{Efficiency and accuracy comparison on E-VQA. Runtime costs are normalized to zero-shot generation; accuracy is reported in percentage points. * denotes a fine-tuned answer generator under the same setting as mKG-RAG\textsuperscript{*}.}
\label{tab:efficiency}
\setlength{\tabcolsep}{2pt}
\resizebox{\linewidth}{!}{
\begin{tabular}{ccccc|c}
\toprule
\textbf{Model}      
& \makecell[c]{\textbf{Document} \\ \textbf{Retrieval}}
& \makecell[c]{\textbf{Graph} \\ \textbf{Retrieval}}
& \makecell[c]{\textbf{Answer} \\ \textbf{Generation}}
& \makecell[c]{\textbf{Total} \\ \textbf{Cost}}
& \textbf{Accuracy}  \\
\midrule
LLaVA-MORE                      & --             & --           & 1.00        & 1.00     & 16.0 \\
Naive RAG\textsuperscript{*}   & 2.29           & --           & 2.20        & 4.49     & 26.5 \\
mKG-RAG\textsuperscript{*}     & 1.87           & 1.12         & 2.13        & 5.12     & 36.3 \\
\bottomrule
\end{tabular}
}
\end{table}
\subsection{Ablation Study}
\subsubsection{\textbf{Impact of Document Retrieval}} 
To quantify the impact of coarse-grained document retrieval, we conduct an ablation experiment replacing the QM-Retriever with visual-only CLIP (ViT-L/14@336) for top-$K_d$ document selection. 
The results in Table~\ref{tab:ablation} reveal significant performance drops: overall VQA accuracy of mKG-RAG decreases by 4.7\% on E-VQA and 2.1\% on InfoSeek. 
This ablation demonstrates the critical role of first-stage retrieval and the superiority of our QM-Retriever over unimodal alternatives.

\subsubsection{\textbf{Effectiveness of Graph Retrieval}}
In our method, the entities and relations extracted from documents form a distilled knowledge graph, reducing noise and enabling more effective retrieval than direct text chunk matching.
To validate this insight, we replace graph-based retrieval with a naive chunk-based alternative. Specifically, we segment retrieved documents into fixed-size chunks and select the relevant ones based on the given question and image caption.
In Table~\ref{tab:ablation}, chunk-based retrieval results in a substantial accuracy drop of 8.1\% on E-VQA and 7.5\% on InfoSeek.

\begin{table*}[t]
\centering
\caption{VQA performance on E-VQA across different MLLM architectures with varying parameter sizes.}
\label{tab:mllm-architecture}
\resizebox{\linewidth}{!}{
\begin{tabular}{cccccccccccc}
\toprule
\multirow{2}{*}{\textbf{MLLM}}      & \multirow{2}{*}{\textbf{E-VQA}} & \textbf{Phi3V}      & \textbf{InternVL3}   & \textbf{LLaMA-3.2}   & \multicolumn{2}{c}{\textbf{LLaVA-v1.5}} & \multicolumn{2}{c}{\textbf{DeepSeek-VL2}} & \multicolumn{3}{c}{\textbf{Qwen2.5-VL}}       \\
\cmidrule(lr){3-12}
                           &                        & 4B         & 8B          & 11B         & 7B            & 13B            & 3B              & 16B            & 3B         & 7B         & 32B        \\
\midrule
\multirow{2}{*}{Zero-shot} & Single-Hop             & 17.7       & 22.4        & 27.0        & 15.8          & 16.1           & 22.0            & 22.4           & 19.1       & 21.0       & 27.1       \\
                           & All                    & 18.0       & 23.0        & 28.9        & 16.2          & 16.6           & 21.6            & 22.3           & 18.9       & 20.8       & 27.3       \\
\midrule
\multirow{2}{*}{mKG-RAG}   & Single-Hop             
& \makecell[c]{26.9 \textcolor{red}{$\uparrow$9.2}} 
& \makecell[c]{32.7 \textcolor{red}{$\uparrow$10.3}} 
& \makecell[c]{37.2 \textcolor{red}{$\uparrow$10.2}} 
& \makecell[c]{25.0 \textcolor{red}{$\uparrow$9.2}}     
& \makecell[c]{27.7 \textcolor{red}{$\uparrow$11.6}}    
& \makecell[c]{28.4 \textcolor{red}{$\uparrow$6.4}}      
& \makecell[c]{31.1 \textcolor{red}{$\uparrow$8.7}}     
& \makecell[c]{28.9 \textcolor{red}{$\uparrow$9.8}} 
& \makecell[c]{30.4 \textcolor{red}{$\uparrow$9.4}} 
& \makecell[c]{36.5 \textcolor{red}{$\uparrow$9.4}} \\
                          & All                    
& \makecell[c]{25.8 \textcolor{red}{$\uparrow$7.8}} 
& \makecell[c]{32.7 \textcolor{red}{$\uparrow$9.7}}  
& \makecell[c]{38.5 \textcolor{red}{$\uparrow$9.6}}
& \makecell[c]{24.6 \textcolor{red}{$\uparrow$8.4}}    
& \makecell[c]{27.8 \textcolor{red}{$\uparrow$11.2}}    
& \makecell[c]{27.4 \textcolor{red}{$\uparrow$5.8}}      
& \makecell[c]{29.9 \textcolor{red}{$\uparrow$7.6}}     
& \makecell[c]{28.2 \textcolor{red}{$\uparrow$9.3}} 
& \makecell[c]{29.6 \textcolor{red}{$\uparrow$8.8}} 
& \makecell[c]{36.5 \textcolor{red}{$\uparrow$9.2}} \\
\bottomrule
\end{tabular}
}
\end{table*}

\begin{table}[t]
\caption{The ablation study on the design of mKG-RAG.}
\label{tab:ablation}
\setlength{\tabcolsep}{2pt}
\resizebox{\linewidth}{!}{
\begin{tabular}{lccccc}
\toprule
\multirow{2}{*}{\textbf{Method}} 
& \multicolumn{2}{c}{\textbf{E-VQA}}
& \multicolumn{3}{c}{\textbf{InfoSeek}} \\
\cmidrule(lr){2-3} \cmidrule(lr){4-6}
                            & Single-Hop    & All           & Unseen-Q          & Unseen-E          & All  \\
\midrule
mKG-RAG\textsuperscript{*}  & \best{38.4} & \best{36.3} & \best{41.4} & \best{39.6} & \best{40.5} \\
w/o QM-Retriever            & 34.2          & 31.6          & 38.9          & 37.9          & 38.4 \\
w/o Graph Retrieval         & 30.1          & 28.2          & 33.3          & 32.7          & 33.0 \\
w/o Subgraph Expansion         & 37.2          & 35.0          & 40.8          & 39.4          & 40.1 \\
\bottomrule
\end{tabular}
}
\end{table}

\subsubsection{\textbf{Contribution of Subgraph Expansion}}
mKG-RAG enhances the constructed subgraph through $l$-hop neighbor expansion, effectively capturing potentially missing but relevant knowledge connections. 
Table~\ref{tab:ablation} shows that omitting graph expansion leads to consistent performance drops of 1.3\% (E-VQA) and 0.4\% (InfoSeek), demonstrating its positive contribution to our approach.

 \begin{table}[t]
\caption{Effect of the number of retrieved entities/relations on VQA accuracy on E-VQA.}
\label{tab:ablation-topk}
\resizebox{\linewidth}{!}{
\begin{tabular}{cccccc}
\toprule
\textbf{Model} & \textbf{Ret. Mode} & \textbf{$K_g=1$} & \textbf{$K_g=5$} & \textbf{$K_g=10$} & \textbf{$K_g=20$} \\
\midrule
mKG-RAG\textsuperscript{*}\dag  & Textual        & 29.1 & 33.9 & 34.9 & 35.9 \\
mKG-RAG\textsuperscript{*}\ddag & Visual         & 23.0 & 29.6 & 31.0 & 32.0 \\
mKG-RAG\textsuperscript{*}      & Multimodal     & \best{29.2} & \best{35.1} & \best{36.3} & \best{36.9} \\
\bottomrule
\end{tabular}
}
\end{table}
 
\subsubsection{\textbf{Impact of Varying Retrieval Number}}
In Table~\ref{tab:ablation-topk}, we further analyze the impact of $K_g$, the number of retrieved entities and relations, on our method.
As $K_g$ increases from 1 to 20, the overall accuracy of mKG-RAG and its variants gradually improves, as higher recall rates enhance the likelihood of capturing relevant knowledge. 
However, when $K_g > 10$, the benefit diminishes due to longer contexts and more noise. Thus, setting $K_g = 10$ offers a practical trade-off.
Notably, mKG-RAG still performs competitively even at $K_g = 1$, thanks to its subgraph expansion strategy, which enables the model to gather additional relevant information.

\subsubsection{\textbf{Consistency across Architectures}}
In Table~\ref{tab:mllm-architecture}, we provide a detailed comparison of VQA scores across different MLLMs with varying parameter sizes, including Phi3V~\cite{marah2024phi3}, InternVL3~\cite{zhu2025internvl3}, LLaMA-3.2-Vision~\cite{llama-3.2-vision}, LLaVA-v1.5~\cite{liu2024improved}, DeepSeek-VL2~\cite{wu2024deepseek}, and Qwen2.5-VL~\cite{bai2025qwen25vltechnicalreport}.
When enhanced with our mKG-RAG framework, these models achieve average performance gains of 9.4\% on single-hop queries and 8.7\% on the overall set, demonstrating the method's strong generalization across different architectures and scales.

\newtcolorbox{mybox}[2][]{%
  enhanced,
  width=\linewidth,
  boxrule=0.5pt,
  arc=2mm,
  colback=gray!10,
  colframe=gray!50,
  coltitle=white,
  colbacktitle=gray!70,
  title=#2,
  #1
}

\begin{figure*}[htbp]
\begin{mybox}{Vision-Text Matching Prompt}
Based on the input image, visual scene graph, and textual graph, please match visual objects and relations in the image with the corresponding entities and relations in the textual graph.\\

\textbf{Input Format}:\\
Each textual entity is formatted as (``entity''\textbar\verb|<entity-name>|\textbar\verb|<entity-type>|\textbar\verb|<entity-description>|), which contains the following information:
(1) \verb|entity-name|: Name of the entity;
(2) \verb|entity-type|: Name of the entity type;
(3) \verb|entity-description|: Comprehensive description of the entity's attributes and activities.\\

Each textual relation is formatted as (``relation''\textbar\verb|<source-entity>|\textbar\verb|<target-entity>|\textbar
\verb|<relation-description>|\\\textbar\verb|<relation-strength>|), which contains the following information:
(1) \verb|source-entity|: name of the source entity, as defined in the textual entities;
(2) \verb|target-entity|: name of the target entity, as defined in the textual entities;
(3) \verb|relation-description|: explanation as to why the source entity and the target entity are related to each other;
(4) \verb|relation-strength|: a numeric score indicating the strength of the relation between the source and target entities, ranging from 0 to 10.\\

Visual scene graph provides objects and relations in the image, formatted as: 
\verb|<object-0>|: \verb|<object-category>|, \verb|<object-bbox>|; 
\verb|<object-1>|: \verb|<object-category>|, \verb|<object-bbox>|;
...;
\verb|<relation-0>|: \verb|<object-0>| \verb|<relation-name>| \verb|<object-1>|
...
The \verb|<object-bbox>| is the bounding box of each object region, represented as (x1, y1, x2, y2) with floating numbers ranging from 0 to 1. These values correspond to the top-left x, top-left y, bottom-right x, and bottom-right y.\\

\textbf{Matching Steps}:\\
\textit{Step 1}. Identify the textual entity that is most relevant to the overall image and extract the following information: 
(1) \verb|entity-name|: the name of the entity that best represents the overall image;
(2) \verb|strength|: a numeric score indicating the strength of the match, ranging from 0 to 10. 
Format the image matching as (``matching''\textbar\verb|<image>|\textbar\verb|<entity-name>|\textbar\verb|<strength>|)

\textit{Step 2}. For each object in the scene graph, if the object visually depicts a textual entity identified in the input data, extract the following information: 
(1) \verb|object-id|: the ID of the object in the scene graph; 
(2) \verb|entity-name|: the name of the entity it represents;
(3) \verb|strength|: a numeric score indicating the strength of the match, ranging from 0 to 10.
Format each object matching as (``matching''\textbar\verb|<object-id>|\textbar\verb|<entity-name>|\textbar\verb|<strength>|)

\textit{Step 3}. For each relation in the scene graph, if the relation visually represents a textual relation identified in the input data, extract the following information:
(1) \verb|relation-id|: the id of the relation in the scene graph;
(2) \verb|source-entity|: the source entity of the relation it represents;
(3) \verb|target-entity|: the target entity of the relation it represents;
(4) \verb|strength|: a numeric score indicating the strength of the match, ranging from 0 to 10. 
Format each relation matching as (``matching''\textbar\verb|<relation-id>|\textbar\verb|<source-entity>|\textbar\verb|<target-entity>|\textbar
\verb|<strength>|)

\textit{Step 4}. For those objects or relations without a corresponding text entity or relation, please ignore them.\\

\textbf{Real Input}: \\
<IMAGE> [Textual Entities \& Relations]
[Visual Entities \& Relations]
\end{mybox}

\caption{The prompt used to match visual and textual entities/relations.}
\label{fig:vision-text-prompt-mini}
\end{figure*}

\section{Conclusion}
This paper proposes mKG-RAG, a novel retrieval-augmented generation framework that integrates multimodal KGs to address the knowledge limitations of MLLMs in knowledge-intensive VQA.
The framework constructs well-structured, modality-complementary multimodal KGs through MLLM-driven graph extraction and cross-modal alignment.
To effectively leverage these KGs, we further introduced a dual-stage multimodal retrieval scheme that combines efficient document-level recall with multi-granularity graph retrieval, allowing precise evidence identification for answer generation.
Extensive experiments demonstrated that mKG-RAG outperforms state-of-the-art methods, with ablation studies validating the contributions of each component.

\begin{acks}

The research described in this paper has been partially supported by the Key Special Project of National Natural Science Foundation of China (Project No. 72442017), the General Research Funds from the Hong Kong Research Grants Council (project No. PolyU 15207322, 15200023, 15206024, and 15224524), Hong Kong Research Grants Council’s Theme-based Research Scheme (No. T43-513/23-N), Hong Kong Research Grants Council’s Research Impact Fund (No. R1015-23), Hong Kong Research Grants Council’s Collaborative Research Fund (No. C1043-24GF), Internal research funds from Hong Kong Polytechnic University (project no. P0059586, P0042693, P0048625, and P0051361), and Sheertek International (HK) Limited. This work was supported by computational resources provided by The Centre for Large AI Models (CLAIM) of The Hong Kong Polytechnic University. 
\end{acks}

\clearpage

\bibliographystyle{ACM-Reference-Format}
\balance
\bibliography{bibliography}


\end{document}